\pdfoutput=1

\documentclass[11pt]{article}

\usepackage[]{ACL2023}

\usepackage{times}
\usepackage{latexsym}

\usepackage[T1]{fontenc}

\usepackage[utf8]{inputenc}

\usepackage{microtype}

\usepackage{inconsolata}

\usepackage{arydshln}
\usepackage{algorithm}
\usepackage[noend]{algpseudocode}
\usepackage{dsfont}
\usepackage{xcolor}
\usepackage{graphicx}
\usepackage{caption}
\usepackage{subcaption}
\usepackage{amsmath}
\usepackage{multirow}

\usepackage{tikz}
\usepackage{etoolbox}
\newcommand*{\lowmark}[1]{\lower.85em \hbox{\tikz\draw (0pt, 0pt)%
    circle (.5em) node {\makebox[0.5em][c]{\tiny #1}};}}
\robustify{\lowmark}
\usepackage{soul}

\title{Target-Side Augmentation for Document-Level Machine Translation}

\author{
    Guangsheng Bao\textsuperscript{\rm 1,2},
    Zhiyang Teng\textsuperscript{\rm 3}, and
    Yue Zhang\thanks{* Corresponding author.} \textsuperscript{\rm ,2,4},
    \\
    \textsuperscript{1} Zhejiang University \\
    \textsuperscript{2} School of Engineering, Westlake University \\
    \textsuperscript{3} Nanyang Technological University \\
    \textsuperscript{4} Institute of Advanced Technology, Westlake Institute for Advanced Study \\
    \textsuperscript{2} \texttt{\{baoguangsheng, zhangyue\}@westlake.edu.cn} \\
    \textsuperscript{3} \texttt{zhiyang.teng@ntu.edu.sg}
}

\begin{document}
\maketitle
\begin{abstract}
Document-level machine translation faces the challenge of data sparsity due to its long input length and a small amount of training data, increasing the risk of learning spurious patterns. To address this challenge, we propose a target-side augmentation method, introducing a data augmentation (DA) model to generate many potential translations for each source document. Learning on these wider range translations, an MT model can learn a smoothed distribution, thereby reducing the risk of data sparsity. We demonstrate that the DA model, which estimates the posterior distribution, largely improves the MT performance, outperforming the previous best system by 2.30 s-BLEU on News and achieving new state-of-the-art on News and Europarl benchmarks. Our code is available at \url{https://github.com/baoguangsheng/target-side-augmentation}.
\end{abstract}
\section{Introduction}
Document-level machine translation \cite{gong2011cache,hardmeier2013docent,werlen2018document,maruf2019selective,bao2021g,feng2022learn} has received increasing research attention. It addresses the limitations of sentence-level MT by considering cross-sentence co-references and discourse information, and therefore can be more useful in the practical setting. Document-level MT presents several unique technical challenges, including significantly longer inputs~\cite{bao2021g} and relatively smaller training data compared to sentence-level MT~\cite{junczys2019microsoft,liu2020multilingual,sun2022rethinking}. The combination of these challenges leads to increased data sparsity~\cite{gao2014learning,koehn2017six,liu2020multilingual}, which raises the risk of learning spurious patterns in the training data~\cite{belkin2019reconciling,savoldi2021gender} and hinders generalization~\cite{li2021compositional,dankers2022paradox}.

\begin{figure*}[t]
    \centering
    \includegraphics[width=1\linewidth]{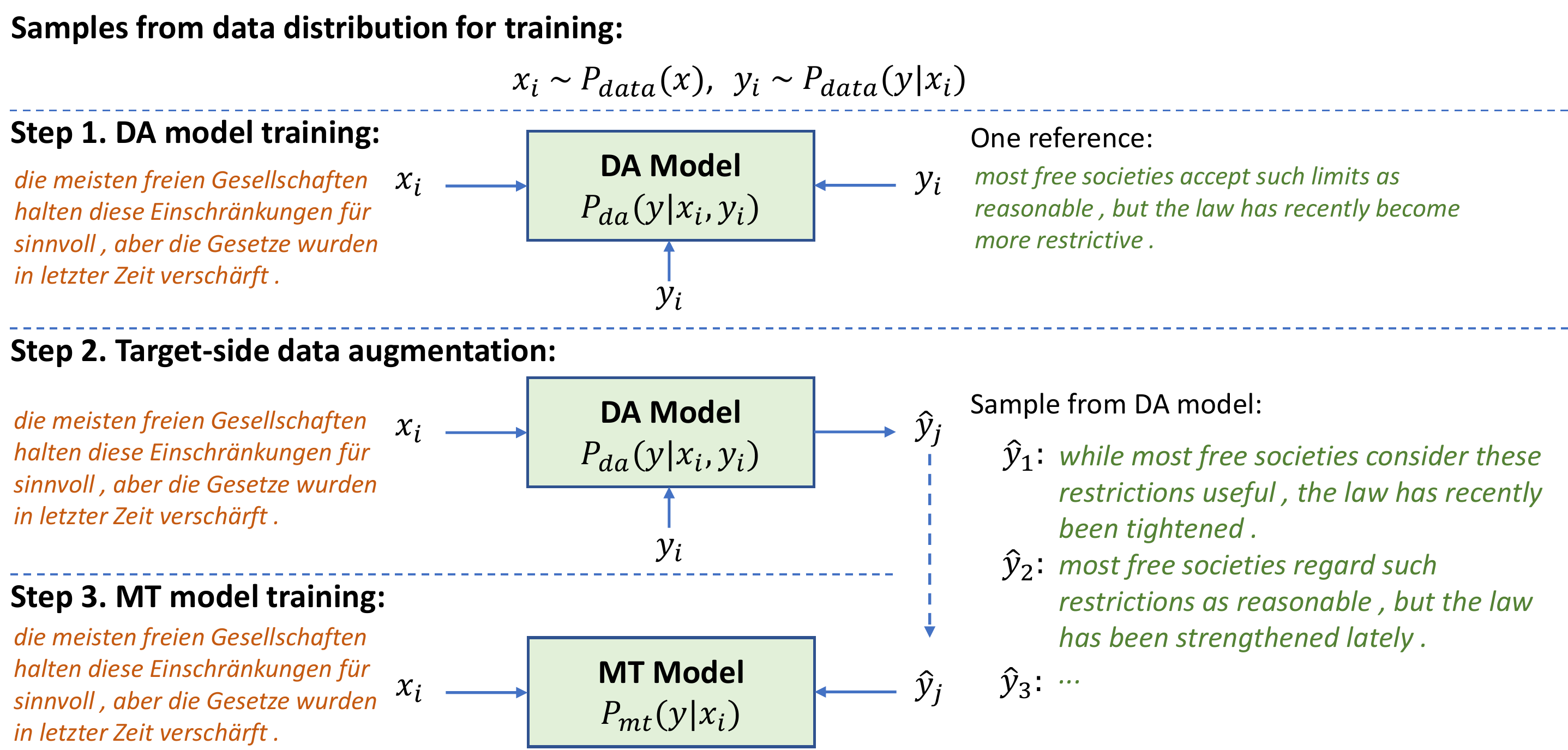}
    \caption{Illustration of target-side data augmentation (DA) using a very simple example. A DA model is trained to estimate the distribution of possible translations $y$ given a source $x_i$ and an observed target $y_i$, and the MT model is trained on the sampled translations $\hat{y}_j$ from the DA model for each source $x_i$. Effectively training the DA model with the target $y_i$, which is also a conditional input, can be challenging, but it is achievable after introducing an intermediate latent variable between the translation $y$ and the condition $y_i$.}
    \label{fig:oracledistillation}
\end{figure*}

To address these issues, we propose a target-side data augmentation method that aims to reduce sparsity by automatically smoothing the training distribution. The main idea is to train the document MT model with many plausible potential translations, rather than forcing it to fit a single human translation for each source document. This allows the model to learn more robust and generalizable patterns, rather than being overly reliant on features of particular training samples. 
Specifically, we introduce a data augmentation (DA) model to generate possible translations to guide MT model training. As shown in Figure~\ref{fig:oracledistillation}, the DA model is trained to understand the relationship between the source and possible translations based on one observed translation (Step 1), and then used to sample a set of potentially plausible translations (Step 2). These translations are fed to the MT model for training, smoothing the distribution of target translations (Step 3).

We use standard document-level MT models including Transformer \cite{vaswani2017attention} and G-Transformer \cite{bao2021g} for both our DA and MT models. For the DA model, in order to effectively capture a \emph{posterior} target distribution given a reference target, we concatenate each source sentence with a latent token sequence as the new input, where the latent tokens are sampled from the observed translation.
A challenge to the DA model is that having the reference translation in the input can potentially decrease diversity. To address this issue, we introduce the intermediate latent variable on the encoder side by using rules to generate n-gram samples, so that posterior sampling \cite{wang2020brief} can be leveraged to yield diverse translations.

Results on three document-level MT benchmarks demonstrate that our method significantly outperforms Transformer and G-Transformer baselines, achieving an improvement of 1.33 and 1.75 s-BLEU on average, respectively, and the state-of-the-art results on News and Europarl. Further analysis shows that high diversity among generated translations and their low deviation from the gold translation are the keys to improved performance.
To our knowledge, we are the first to do \emph{target-side} augmentation to enrich \emph{output} variety for document-level machine translation.

\section{Related Work}
\textbf{Data augmentation (DA)} increases training data by synthesizing new data~\cite{van2001art,shorten2019survey,shorten2021text,li2022data}. In neural machine translation (NMT), the most commonly used data augmentation techniques are \textbf{source-side augmentations}, including easy data augmentation (EDA) \cite{wei2019eda}, subword regularization \cite{kudo2018subword}, and back-translation~\cite{sennrich2016improving}, which generates pseudo sources for monolingual targets enabling the usage of widely available monolingual data. These methods generate more source-target pairs with different silver source sentences for the same gold-target translation.
On the contrary, \textbf{target-side augmentation} is more challenging, as approaches like EDA are not effective for the target side because they corrupt the target sequence, degrading the autoregressive modeling of the target language.

Previous approaches on target-side data augmentation in NMT fall into three categories. The first is based on \emph{self-training}~\cite{bogoychev2019domain,he2019revisiting,zoph2020rethinking}, which generates pseudo translations for monolingual source text using a trained model. The second category uses either a pre-trained language model~\cite{fadaee2017data,wu2019conditional} or a pre-trained generative model~\cite{raffel2020exploring,khayrallah2020simulated} to generate \emph{synonyms} for words or \emph{paraphrases} of the target text. The third category relies on reinforcement learning \cite{norouzi2016reward,wang2018switchout}, introducing a reward function to evaluate the quality of translation candidates and to regularize the likelihood objective. In order to explore possible candidates, a sampling from the model distribution or random noise is used. Unlike these approaches, our method is a target-side data augmentation technique that is trained using supervised learning and does not rely on external data or large-scale pretraining. More importantly, we generate document-level instead of word, phrase, or sentence-level alternatives. 

Previous target-side input augmentation \cite{xie2022target} appears to be similar to our target-side augmentation. However, besides the literal similarity, they are quite different. Consider the token prediction $P(y_i|x, y_{<i})$. The target-side input augmentation augments the condition $y_{<i}$ to increase the model's robustness to the conditions, which is more like source-side augmentation on condition $x$. In comparison, target-side augmentation augments the target $y_i$, providing the model with completely new training targets.

\textbf{Paraphrase models.} 
Our approach generates various translations for each source text, each of which can be viewed as a paraphrase of the target. Unlike previous methods that leverage paraphrase models for improving MT \cite{madnani2007using,hu2019improved,khayrallah2020simulated}, our DA model exploits parallel corpus and does not depend on external paraphrase data, similar to \citet{thompson2020paraphrase}.
Instead, it takes into account the source text when modeling the target distribution. More importantly, while most paraphrase models operate at the sentence level, our DA model can generate translations at the document level.

\begin{figure*}[t]
    \centering
    \includegraphics[width=1\linewidth]{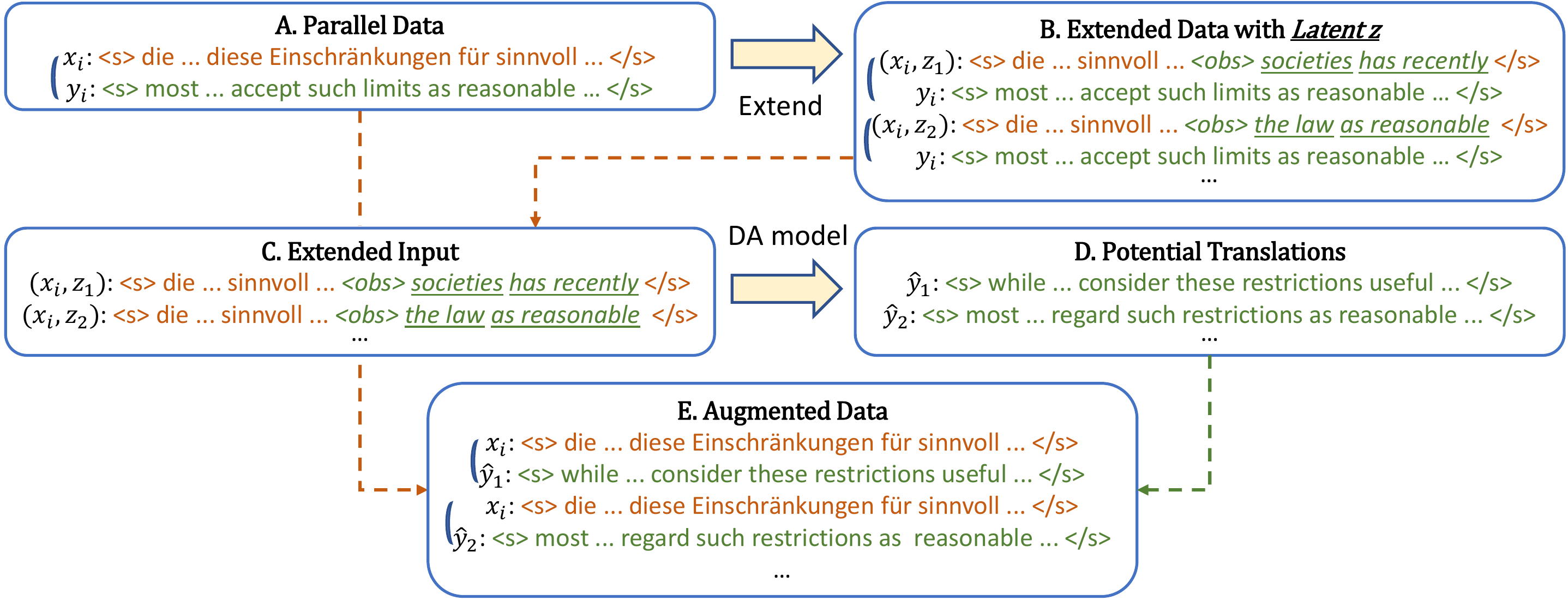}
    \caption{The detailed data augmentation process, where the parallel data is augmented multiple times.}
    \label{fig:training-process}
\end{figure*}

\textbf{Conditional auto-encoder.} 
The DA model can also be seen as a conditional denoising auto-encoder (c-DAE), where the latent variable is a noised version of the ground-truth target, and the model is trained to reconstruct the ground-truth target from a noisy latent sequence. c-DAE is similar to the conditional variational autoencoder (c-VAE)~\cite{zhang2016variational,pagnoni2018conditional}, which learns a latent variable and generates diverse translations by sampling from it. 
However, there are two key differences between c-VAE and our DA model. First, c-VAE learns both the prior and posterior distributions of the latent variable, while the DA model directly uses predefined rules to generate the latent variable. Second, c-VAE models the prior distribution of the target, while the DA model estimates the posterior distribution.

\textbf{Sequence-level knowledge distillation.} 
Our DA-MT process is also remotely similar in form to sequence-level knowledge distillation (SKD) ~\cite{ba2014deep,hinton2015distilling,gou2021knowledge,kim2016sequence,gordon2019explaining,lin2020autoregressive}, which learns the data distribution using a large teacher and distills the knowledge into a small student by training the student using sequences generated by the teacher. 
However, our method differs from SKD in three aspects. 
First, SKD aims to compress knowledge from a large teacher to a small student, while we use the same or smaller size model as the DA model, where the knowledge source is the training data rather than the big teacher. 
Second, the teacher in SKD estimates the prior distribution of the target given source, while our DA model estimates the posterior distribution of the target given source and an observed target. 
Third, SKD generates one sequence for each source, while we generate multiple diverse translations with controlled latent variables.
\section{Target-Side Augmentation}
The overall framework is shown in Figure \ref{fig:oracledistillation}. 
Formally, denote a set of training data as $D=\{(x_i,y_i)\}_{i=1}^{N}$, where $(x_i, y_i)$ is the $i$-th source-target pair and $N$ is the number of pairs. We train a data augmentation (DA) model (Section \ref{subsec:da-model}) to generate samples with new target translations (Section \ref{subsec:da-process}), which are used to train an MT model (Section \ref{subsec:mt-model}).

\subsection{The Data Augmentation Model}
\label{subsec:da-model}
We learn the posterior distribution $P_{da}(y|x_i,y_i)$ from parallel corpus by introducing latent  variables
\begin{equation}
  P_{da}(y|x_i,y_i) = \sum_{z\in \mathcal{Z}_i} P_{\varphi}(y|x_i,z) P_{\alpha}(z|y_i), 
  \label{eq:posterior}
\end{equation}
where  $z$ is the latent variable to control the translation output and $\mathcal{Z}_i$ denotes the possible space of $z$, $\varphi$ denotes the parameters of the DA model, and $\alpha$ denotes the hyper-parameters for determining the distribution of $z$ given $y_i$. 

The space $\mathcal{Z}_i$ of possible $z$ is exponentially large compared to the number of tokens of the target, making it intractable to sum over $\mathcal{Z}_i$ in Eq. \ref{eq:posterior}. We thus consider a Monte Carlo approximation, sample a group of instances from $p_{\alpha}(z|y_i)$, and calculate the sample mean
\begin{equation}
  P_{da}(y|x_i,y_i) \approx \frac{1}{|\mathcal{\hat{Z}}_i|} \sum_{z\in \mathcal{\hat{Z}}_i} P_{\varphi}(y|x_i,z), 
  \label{eq:posterior-approx}
\end{equation}
where $\mathcal{\hat{Z}}_i$ denotes the sampled instances.

There are many possible choices for the latent variable, such as a continuous vector or a categorical discrete variable, which also could be either learned by the model or predefined by rules. Here, we simply represent the latent variable as a sequence of tokens and use predefined rules to generate the sequence, so that the latent variable can be easily incorporated into the input of a seq2seq model without the need for additional parameters.

Specifically, we set the value of the latent variable $z$ to be a group of sampled n-grams from the observed translation $y_i$ and concatenate $x_i$ and $z$ into a sequence of tokens. We assume that the generated translations $y$ can be consistent with the observed translation $y_i$ on these n-grams.
To this end, we define $\alpha$  as the ratio of tokens in $y_i$ that is observable through $z$, naming \emph{observed ratio}. For a target with $|y_i|$ tokens, we uniformly sample n-grams from $y_i$ to cover $\alpha\times |y_i|$ tokens that each n-gram has a random length among $\{1,2,3\}$. For example, given that $\alpha=0.1$ and a target $y_i$ with $20$ tokens, we can sample one 2-gram or two uni-grams from the target to reach 2 ($0.1\times 20$) tokens.

\textbf{Training.}
Given a sample $(x_i, y_i)$, the training loss is rewritten as
\begin{equation}
\small
\begin{split}
  \mathcal{L}_{da} &= - \sum_{i=1}^{N} \log P_{da}(y=y_i|x_i,y_i)  \\
  &\approx - \sum_{i=1}^{N} \log \frac{1}{|\mathcal{\hat{Z}}_i|} \sum_{z\in \mathcal{\hat{Z}}_i} P_{\varphi}(y=y_i|x_i,z) \\
  &\leq - \sum_{i=1}^{N} \frac{1}{|\mathcal{\hat{Z}}_i|} \sum_{z\in \mathcal{\hat{Z}}_i} \log P_{\varphi}(y=y_i|x_i,z), \\
\end{split}
\label{eq:lossdawithz}
\end{equation}
where the upper bound of the loss is provided by Jensen inequality. The upper bound sums log probabilities, which can be seen as sums of the standard negative log-likelihood (NLL) loss of each $(x_i, z, y_i)$. As a result, when we optimize this upper bound as an alternative to optimizing $\mathcal{L}_{da}$, the DA model is trained using standard NLL loss but with $|\mathcal{\hat{Z}}_i|$ times more training instances.

\textbf{Discussion.}
As shown in Figure~\ref{fig:oracledistillation}, given a sample $(x_i, y_i)$, we adopt a new estimation method using the posterior distribution $P_{da}(y|x_i,y_i)$ for our DA model. The basic intuition is that by conditioning on both the source $x_i$ and the observed translation $y_i$, the DA model can estimate the data distribution $P_{data}(y|x_i)$ more accurately than an MT model. Logically, an MT model learns a prior distribution $P_{mt}(y|x_i)$, which estimates the data distribution $P_{data}(y|x_i)$ for modeling translation probabilities. This prior distribution works well when the corpus is large. However, when the corpus is sparse in comparison to the data space, the learned distribution overfits the sparsely distributed samples, resulting in poor generalization to unseen targets.

\subsection{The Data Augmentation Process}
\label{subsec:da-process}
The detailed data augmentation process is shown in Figure \ref{fig:training-process} and the corresponding algorithm is shown in Algorithm~\ref{alg:target-aug}. Below we use one training example to illustrate.

\textbf{DA model training.}
We represent the latent variable $z$ as a sequence of tokens and concatenate $z$ to the source, so a general seq2seq model can be used to model the posterior distribution. Compared to general MT models, the only difference is the structure of the input.

Specifically, as the step B shown in the figure,  for a given sample $(x_i,y_i)$ from the parallel data, we sample a number of n-grams from $y_i$ and extend the input to $(x_i,z)$, where the number is determined according to the length of $y_i$. Take the target sentence ``{\it most free societies accept such limits as reasonable , but the law has recently become more restrictive .}'' as an example. We sample ``{\it societies}'' and ``{\it has recently}'' from the target and concatenate them to the end of the source sentence to form the first input sequence. We then sample ``{\it the law}'' and ``{\it as reasonable}'' to form the second input sequence.  These new input sequences pair with the original target sequence to form new parallel data. By generating different input sequences, we augment the data multiple times.

\begin{algorithm}[t]
\small
\hspace*{\algorithmicindent} {\bf Input}: $D=\{(x_i, y_i)\}_{i=1}^{N}$ \; \Comment{A. Parallel data} \\
\hspace*{\algorithmicindent} {\bf Output}: $D'=\{(x_i,y_i)\}_{i=1}^{N\times (M+1)}$\; \Comment{Aug $M$ times}
\begin{algorithmic}[1]
\Function{TargetAug}{$D$} 
  \State $D' \gets \{\}$ 
  \For{$i \gets 1$ to $N$}
    \State $(x_i, y_i) \gets D[i]$ \Comment{For each sample}
    \State $D' \gets D'\cup \{(x_i, y_i)\}$ \Comment{Add the gold pair}
    \For{$j \gets 1$ to $M$}
      \State $\alpha \sim Beta(a,b)$  \Comment{Sample an observed ratio}
      \State $z_j \sim P_{\alpha}(z|y_i)$  \Comment{Sample a latent value}
      \State $\hat{y}_j \sim P_{\varphi}(y|x_i, z_j)$   \Comment{Sample a translation}
      \State $D' \gets D'\cup \{(x_i, \hat{y}_j)\}$    \Comment{Add the new pair}
    \EndFor
  \EndFor
  \State \textbf{return} $D'$  \Comment{E. Augmented data}
\EndFunction 
\end{algorithmic}
\caption{Target-side data augmentation.}
\label{alg:target-aug}
\end{algorithm}

\textbf{Target-side data augmentation.}
Using the data ``C. Extended Input'' separated from the extended data in step B, we generate new translations by running a beam search with the trained DA model, where for each extended input sequence, we obtain a new translation. Here, we reuse the sampled $z$ from step B. However, we can also sample new $z$ for inference, which does not show an obvious difference in the MT performance.
By pairing the new translations with the original source sequence, we obtain ``E. Augmented Data''. The details are described in Algorithm \ref{alg:target-aug}, which inputs the original parallel data and outputs the augmented data.

\subsection{The MT Model}
\label{subsec:mt-model}
We use Transformer~\cite{vaswani2017attention} and G-Transformer~\cite{bao2021g} as the baseline MT models. The Transformer baseline models the sentence-level translation and translates a document sentence-by-sentence, while the G-Transformer models the whole document translation and directly translates a source document into the corresponding target document. G-transformer improves the na\"ive self-attention in Transformer with group-attention (Appendix~\ref{sec:gtrans}) for long document modeling, which is a recent state-of-the-art document MT model. 

\textbf{Baseline Training.}
The baseline methods are trained on the original training dataset $D$ by the standard NLL loss
\begin{equation}
  \mathcal{L}_{mt} = - \sum_{i=1}^{N} \log P_{mt}(y=y_i|x_i). \\
\end{equation}

\textbf{Augmentation Training.}
For our target-side augmentation method, we force the MT model to match the posterior distribution estimated by the DA model
\begin{equation}
  \mathcal{L}_{mt} = - \sum_{i=1}^{N} \sum_{y \in \mathcal{Y}_i} P_{da}(y|x_i,y_i) \log P_{mt}(y|x_i), \\
  \label{eq:loss-student}
\end{equation}
where $\mathcal{Y}_i$ is the possible translations of $x_i$.

We approximate the expectation over $\mathcal{Y}_i$ using a Monte Carlo method. Specifically, for each sample $(x_i, y_i)$, we first sample $z_j$ from $P_{\alpha}(z|y_i)$ and then run beam search with the DA model by taking $x_i$ and $z_j$ as its input, obtaining a feasible translation. Repeating the process $M$ times, we obtain a set of possible translations 
\begin{equation}
  \mathcal{\hat{Y}}_i = \{\arg \max_{y}  P_{\varphi}(y|x_i,z_j)| z_j \sim P_{\alpha}(z|y_i)\}_{j=1}^{M}, \\
\end{equation}
as the step D in Figure \ref{fig:training-process} and Algorithm \ref{alg:target-aug} in Section \ref{subsec:da-process} illustrate.

Subsequently, the loss function for the MT model is rewritten as follows, which approximates the expectation using the average NLL loss of the sampled translations
\begin{equation}
  \mathcal{L}_{mt} \approx  - \sum_{i=1}^{N} \frac{1}{|\mathcal{\hat{Y}}_i|} \sum_{y \in \mathcal{\hat{Y}}_i}  \log P_{\theta}(y|x_i), \\
\label{eq:loss-student-approx}
\end{equation}
where $\theta$ denotes the parameters of the MT model. The number $|\mathcal{\hat{Y}}_i|$ could be different for each sample, but for simplicity, we choose a fixed number $M$ in our experiments.

\begin{table}[t]
    \small
    \centering
    \begin{tabular}{lcc}
        \hline
        \multirow{2}{*}{\bf{Dataset}} & \bf{Sentences} & \bf{Documents}  \\
         & \bf{train/dev/test} & \bf{train/dev/test} \\
        \hline
        TED & 0.21M/9K/2.3K  &  1.7K/92/22  \\
        News & 0.24M/2K/3K  &  6K/80/154  \\
        Europarl & 1.67M/3.6K/5.1K  & 118K/239/359  \\
        \hline
    \end{tabular}
    \caption{Datasets statistics.}
    \label{tab:datasets}
\end{table}

\begin{table*}[t]
    \small
    \centering
    \begin{tabular}{l|c@{\hskip 8pt}cc@{\hskip 8pt}cc@{\hskip 8pt}c|c}
        \hline
        \multirow{2}{*}{\bf Method} & \multicolumn{2}{c}{\bf TED} & \multicolumn{2}{c}{\bf News} & \multicolumn{2}{c|}{\bf Europarl} & {\bf Average} \\
         & s-BLEU & d-BLEU & s-BLEU & d-BLEU & s-BLEU & d-BLEU & s-BLEU \\
        \hline
        HAN \cite{miculicich2018document} & 24.58 & - & 25.03 & - & 28.60 & - &  26.07  \\
        SAN \cite{maruf2019selective} & 24.42 & - & 24.84 & - & 29.75 & - & 26.34 \\
        Hybrid Context \cite{zheng2020toward} & 25.10 & - & 24.91 & - & 30.40 & - & 26.80 \\
        Flat-Transformer \cite{ma2020simple} & 24.87 & - & 23.55 & - & 30.09 & - &  26.17 \\
        G-Transformer (rnd.) \cite{bao2021g} & 23.53 & 25.84 & 23.55 & 25.23 & 32.18 & 33.87 & 26.42 \\
        G-Transformer (fnt.) \cite{bao2021g} & 25.12 & 27.17 & 25.52 & 27.11 & 32.39 & 34.08 & 27.68 \\
        MultiResolution \cite{sun2022rethinking} & 25.24 & 29.27 & 25.00 & 26.71 & 32.11 & 34.48 & 27.45 \\
        RecurrentMem \cite{feng2022learn} & 25.62 & {\bf 29.47} & 25.73 & 27.78 & 31.41 & 33.50 & 27.59 \\
        SMDT \cite{zhang2022smdt} & 25.12 & - & 25.76 & - & 32.42 & - & 27.77  \\
        \hline
        Transformer (sent baseline) $\diamondsuit$  & 24.91 & - & 24.82 & - & 31.22 & - & 26.98 \\
        + Target-side data augmentation (ours) & 26.14* & - & 27.03* & - & 31.75* & - & 28.31 \\
        \hdashline
        G-Transformer (fnt.) (doc baseline) $\diamondsuit$ & 25.20 & 27.94 & 25.12 & 27.02 & 31.93 & 33.88 & 27.42 \\
        + Target-side augmentation (ours) & {\bf 26.59}* & 29.20* & 28.06* & 29.83* & {\bf 32.85}* & {\bf 34.76}* & {\bf 29.17} \\
        \hline
        Transformer + Back-translation (sent) $\heartsuit$ & 25.03 & - & 26.07 & - & 31.12 & - & 27.41 \\
        Target-side augmentation (ours) & 26.13 & - & 28.01 & - & 31.27 & - & 28.47 \\
        \hdashline       
        G-Transformer + Back-translation (doc) $\heartsuit$ & 25.45 & 28.06 & 26.25 & 28.21 & 32.00 & 33.94 & 27.90 \\
        Target-side augmentation (ours) & 26.21 & 28.58 & {\bf 28.69} & {\bf 30.41} & 32.52 & 34.50 &  29.14 \\
        \hline          
        \multicolumn{7}{c}{\bf Pre-training Setting for Comparison} \\
        \hline
        Flat-Transformer+BERT \cite{ma2020simple} & 26.61 & - & 24.52 & - & 31.99 & - & 27.71 \\
        G-Transformer+BERT \cite{bao2021g} & 26.81 & - & 26.14 & - & 32.46 & - & 28.47 \\
        G-Transformer+mBART \cite{bao2021g} & 28.06 & 30.03 & 30.34 & 31.71 & 32.74 & 34.31 & 30.38 \\
        \hline
    \end{tabular}
    \caption{Main results evaluated on English-German document-level translation, where ``*'' indicates a significant improvement upon the baseline with $p<0.01$. (rnd.) -- parameters are randomly initialized. (fnt.) -- parameters are initialized using a trained sentence model. $\diamondsuit$ -- we adjust the hyper-parameters for augmented datasets. $\heartsuit$ -- we augment the training data by back-translating each target to a new source instead of introducing additional monolingual targets.}
    \label{tab:main-results}
\end{table*}
\section{Experiments}
\textbf{Datasets.}
We experiment on three benchmark datasets -- TED, News, and Europarl \cite{maruf2019selective}, representing different domains and data scales for English-German (En-De) translation.
The detailed statistics are displayed in Table \ref{tab:datasets}, and the detailed descriptions are in Appendix \ref{subsec:datasets}.

\textbf{Metrics.} 
We follow \citet{liu2020multilingual} to use sentence-level BLEU score (s-BLEU) and document-level BLEU score (d-BLEU) as the major metrics for the \emph{performance}. We further define two metrics, including Deviation and Diversity, to measure the quality of generated translations from the DA model for \emph{analysis}. The detailed description and definition are in Appendix \ref{subsec:metrics}.

\textbf{Baselines.}
We apply target-side augmentation to two baselines, including sentence-level Transformer \cite{vaswani2017attention} and document-level G-transformer \cite{bao2021g}. We further combine back-translation and target-side augmentation, and apply it to the two baselines.

\textbf{Training Settings.}
For both Transformer and G-Transformer, we generate $M$ new translations (9 for TED and News, and 3 for Europarl) for each sentence and augment the data to its $M+1$ times. 
For back-translation baselines, where the training data have already been doubled, we further augment the data 4 times for TED and News, and 1 for Europarl, so that the total times are still 10 for TED and News, and 4 for Europarl.

We obtain the translations by sampling latent $z$ with an observed ratio from a Beta distribution $Beta(2,3)$ and running a beam search with a beam size of 5. We run each main experiment three times and report the median. More details are described in Appendix \ref{subsec:training-settings}.

\subsection{Main Results}
As shown in Table \ref{tab:main-results}, target-side augmentation significantly improves all the \emph{baselines}.  Particularly, it improves G-Transformer (fnt.) by 1.75 s-BLEU on average over the three benchmarks, where the improvement on News reaches 2.94 s-BLEU. 
With the augmented data generated by the DA model, the gap between G-Transformer (rnd.) and G-Transformer (fnt.) narrows from 1.26 s-BLEU on average to 0.18, suggesting that fine-tuning on sentence MT model might not be necessary when augmented data is used. 
For the Transformer baseline, target-side augmentation enhances the performance by 1.33 s-BLEU on average. These results demonstrate that target-side augmentation can significantly improve the baseline models, especially on small datasets.

Comparing with \emph{previous work}, G-Transformer (fnt.)+Target-side augmentation outperforms the best systems SMDT, which references retrieved similar translations, with a margin of 1.40 s-BLEU on average. It outperforms previous competitive RecurrentMem, which gives the best score on TED, with a margin of 1.58 s-BLEU on average. Compared with MultiResolution, which is also a data augmentation approach that increases the training data by splitting the documents into different resolutions (e.g., 1, 2, 4, 8 sentences per training instance), target-side augmentation obtains higher performance with a margin of 1.72 s-BLEU on average. With target-side augmentation, G-Transformer (fnt.) achieves the best-reported s-BLEU on all three datasets.

Compared to the \emph{pre-training setting}, target-side augmentation with G-Transformer (fnt.) outperforms Flat-Transformer+BERT and G-Transformer+BERT, which are fine-tuned on pre-trained BERT, with margins of 1.46 and 0.70 s-BLEU, respectively, on an average of the three benchmarks, where the margins on News reaches 3.54 and 1.92, respectively. The score on bigger dataset Europarl even excels strong large pre-training G-Transformer+mBART, suggesting the effectiveness of target-side augmentation for both small and large datasets.

\emph{Back-translation} does not enhance the performance on TED and Europarl by an adequate margin, but enhances the performance on News significantly, compared to the Transformer and G-Transformer baselines. Upon the enhanced baselines, target-side augmentation further improves the performance on News to a new level, reaching the highest s/d-BLEU scores of 28.69 and 30.41, respectively.
The results demonstrate that target-side augmentation complements the back-translation technique, where a combination may be the best choice in practice.

\subsection{Posterior vs Prior Distribution}
\label{sec:post-prior}
We first compare the MT performance of using a posterior distribution $P(y|x_i, y_i)$ in the DA model (Eq. \ref{eq:loss-student} in Section \ref{subsec:mt-model}) against using the prior distribution $P(y|x_i)$. As shown in Table \ref{tab:post-prior-performance}, when using a prior-based augmentation, the performance improves by 0.64 s-BLEU on average compared to using the original data.
After replacing the DA model with the posterior distribution, the performance improves by 1.75 s-BLEU on average, which is larger than the improvements obtained by the prior distribution.
The results suggest that using a DA model  (even with a simple prior distribution) to augment the target sequence is effective, and the posterior distribution further gives a significant boost.

\begin{table}[!t]
    \small\centering
    \setlength{\abovecaptionskip}{6pt}
    \setlength{\tabcolsep}{3pt}
    \begin{tabular}{l|cc@{\hskip 8pt}c|c}
        \hline
        {\bf Method} & {\bf TED} & {\bf News} & {\bf Europarl} & {\bf Increase}  \\
        \hline
        G-Transformer (fnt.) & 25.20 & 25.12 & 31.93 & - \\
        + Prior-based aug & 25.69 & 26.34 & 32.16 & +0.64 \\
        + Posterior-based aug & 26.59 & 28.06 & 32.85 & +1.75 \\
        \hline
    \end{tabular}
    \caption{MT performance with prior/posterior-based DA models, evaluated in \emph{s-BLEU}.}
    \label{tab:post-prior-performance}
\end{table}

\begin{table}[t]
    \small\centering
    \setlength{\abovecaptionskip}{6pt}
    \setlength{\belowcaptionskip}{-9pt}
    \setlength{\tabcolsep}{4pt}
    \begin{tabular}{l|cc|c}
        \hline
        {\bf Method} & {\bf Diversity $\uparrow$} & {\bf Deviation $\downarrow$} & {\bf PPL $\downarrow$}  \\
        \hline
        Prior distribution & \bf 78.68 & 76.55 & 8.68 \\
        Posterior distribution & 45.42 & \bf 47.14 & \bf 7.00 \\
        \hline
    \end{tabular}
    \caption{Quality of generated translations and accuracy of the estimated distributions from the DA model, evaluated on {\it News}.}
    \label{tab:post-prior-quality}
\end{table}

\textbf{Generated Translations.}
We evaluate the distribution of generated translations, as shown in Table \ref{tab:post-prior-quality}. Using prior distribution, we obtain translations with higher Diversity than posterior distribution. However, higher Diversity does not necessarily lead to better performance if the generated translations are not consistent with the target distribution. As the Deviation column shows, the translations sampled from the posterior distribution have a much smaller Deviation than that from the prior distribution, which confirms that the DA model estimating posterior distribution can generate translations more similar to the gold target.

\textbf{Accuracy of Estimated Distribution.}
As more direct evidence to support the DA model with a posterior distribution, we evaluate the perplexity (PPL) of the model on a multiple-reference dataset, where a better model is expected to give a lower PPL on the references (Appendix \ref{subsec:estimate-distrib}).
As shown in the column PPL in Table \ref{tab:post-prior-quality}, we obtain an average PPL (per token) of 7.00 for the posterior and 8.68 for the prior distribution, with the former being 19.4\% lower than the latter, confirming our hypothesis that the posterior distribution can estimate the data distribution $P_{data}(y|x_i)$ more accurately.

\begin{figure}[t]
    \centering
    \includegraphics[trim={0 12pt 0 12pt},clip,width=0.7\linewidth]{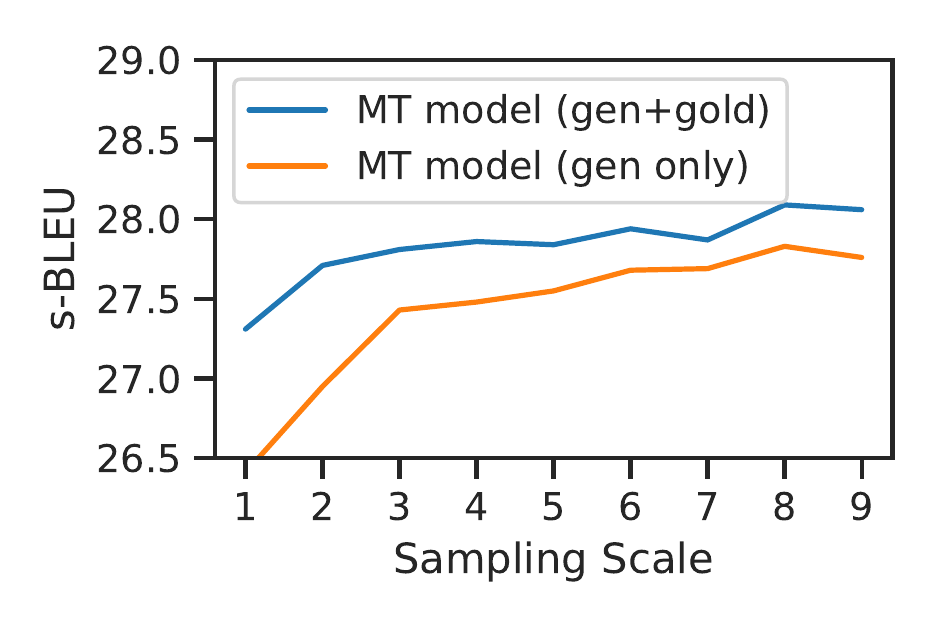}
    \caption{Impact of the sampling scale for $z$, trained on G-Transformer (fnt.) and evaluated in \emph{s-BLEU} on \emph{News}. (gen+gold) -- trained on both generated and gold translations. (gen only) -- trained on generated translations.}
    \label{fig:sampling-scale}
\end{figure}

\begin{figure*}[t]
  \begin{minipage}{0.74\linewidth}
    \centering
    \captionsetup{width=0.98\linewidth}
    \begin{subfigure}[b]{0.328\linewidth}
        \centering
        \captionsetup{width=.9\linewidth}
        \includegraphics[trim={12pt 12pt 12pt 12pt},clip,width=1\linewidth]{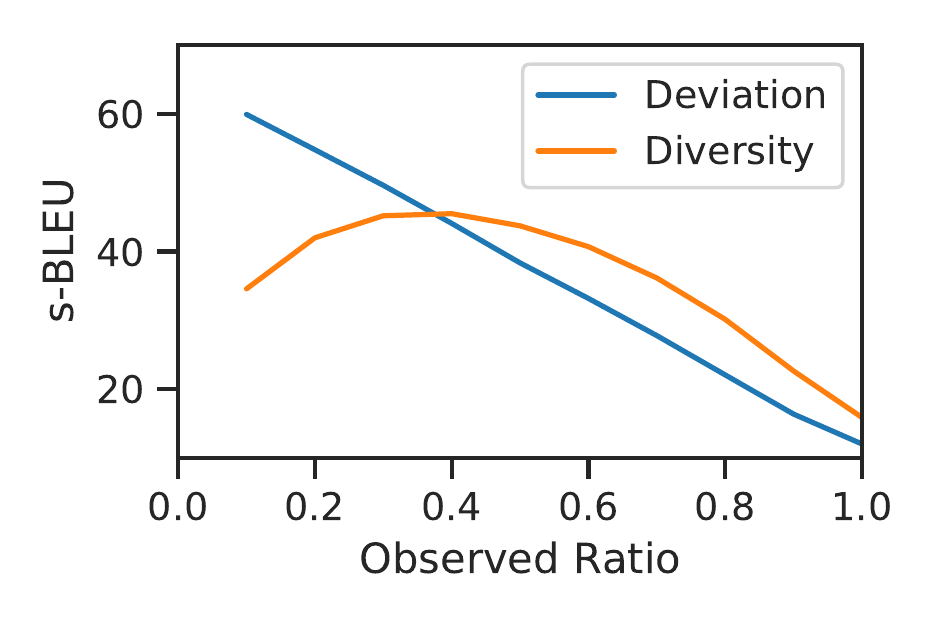}
        \caption{Quality of translations generated by the DA model, evaluated on \emph{News}}
        \label{fig:observe-ratio-a}
    \end{subfigure}
    \begin{subfigure}[b]{0.328\linewidth}
        \centering
        \captionsetup{width=.9\linewidth}
        \includegraphics[trim={12pt 12pt 12pt 12pt},clip,width=1\linewidth]{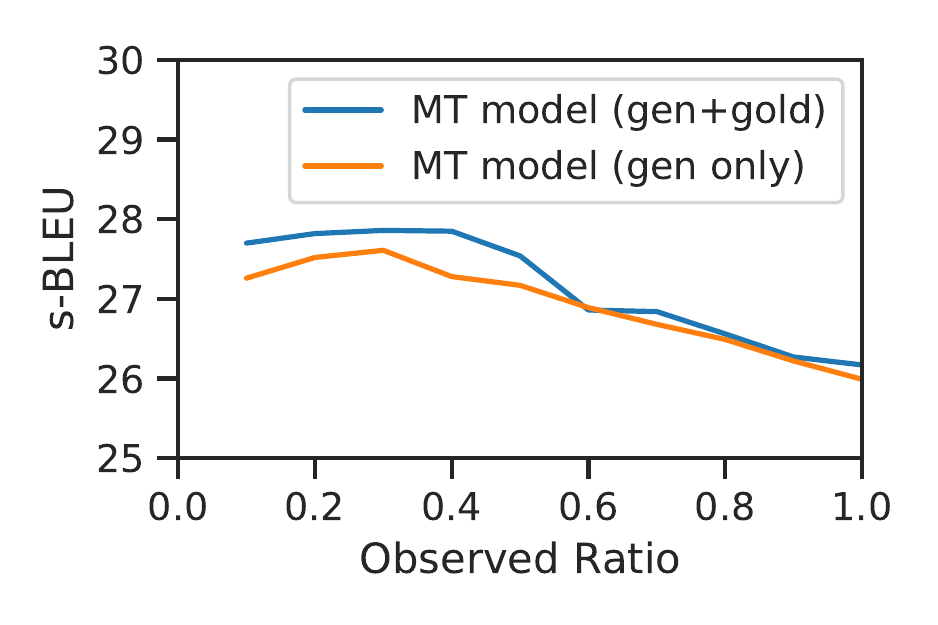}
        \caption{Performance of MT model on augmented data, evaluated on \emph{News}}
        \label{fig:observe-ratio-b}
    \end{subfigure}
    \begin{subfigure}[b]{0.328\linewidth}
        \centering
        \captionsetup{width=.9\linewidth}
        \includegraphics[trim={12pt 12pt 12pt 12pt},clip,width=1\linewidth]{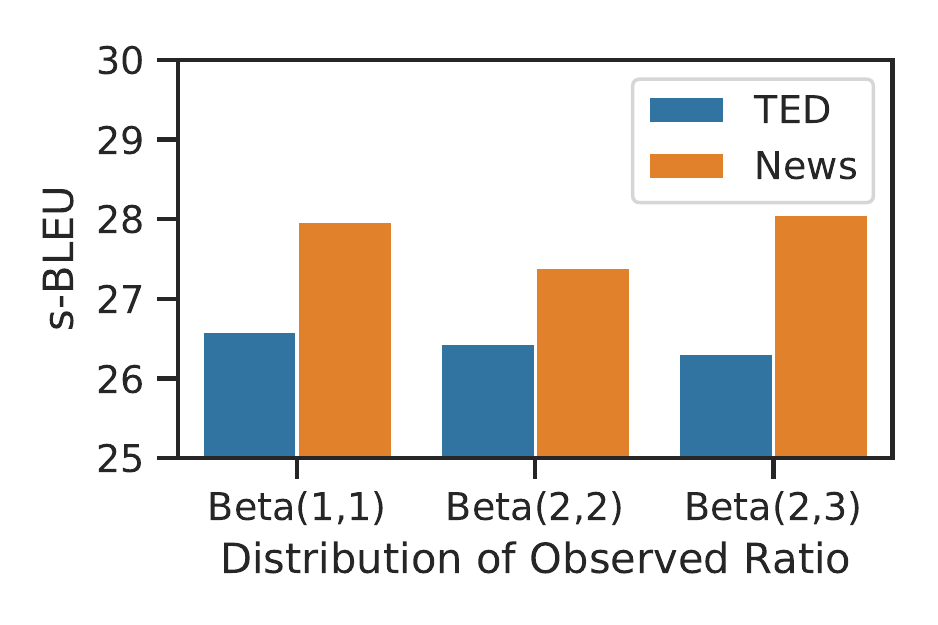}
        \caption{Performance of MT models trained using mixed observed ratios}
        \label{fig:observe-ratio-c}
    \end{subfigure}
    \caption{Impact of the observed ratio for $z$, trained on G-Transformer (fnt.) and evaluated in \emph{s-BLEU}. Beta(a,b) -- The function curves are shown in Appendix \ref{subsec:beta}.}
    \label{fig:observe-ratio}
  \end{minipage}\hfill
  \begin{minipage}{0.246\linewidth}
    \centering
    \captionsetup{width=0.9\linewidth}
    \includegraphics[trim={12pt 12pt 12pt 12pt},clip,width=1\linewidth]{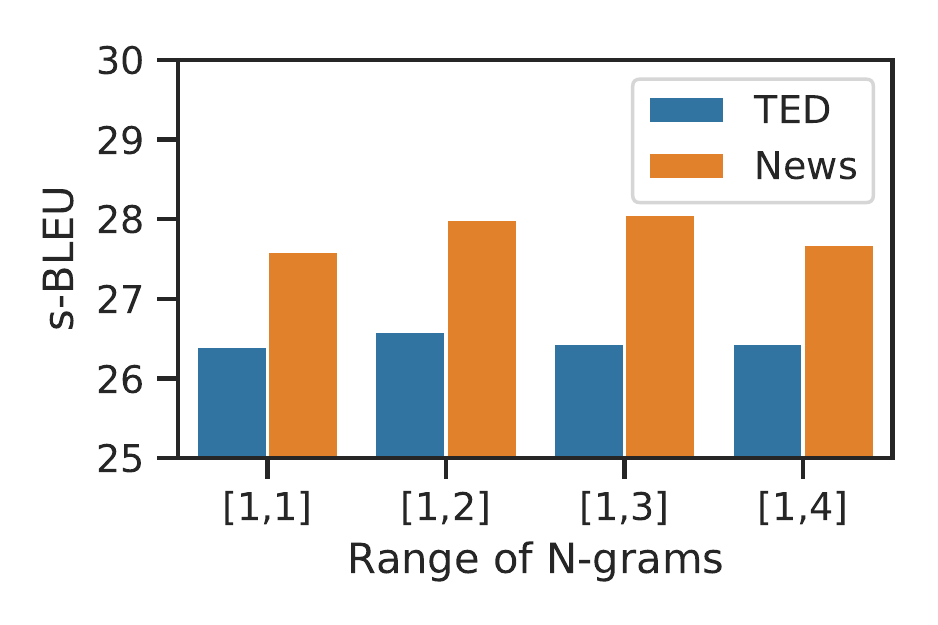}
    \caption{Impact of the granularity of n-grams, trained on G-Transformer (fnt.) and evaluated in \emph{s-BLEU}.}
    \label{fig:observe-ngram}
  \end{minipage}
\end{figure*}

\subsection{Sampling of Latent z}
\label{sec:sampling-latent}

\textbf{Scale.}
The sampling scale $|\mathcal{\hat{Y}}|$ in Eq. \ref{eq:loss-student-approx} is an important influence factor on the model performance. Theoretically, the larger the scale is, the more accurate the approximation will be. Figure \ref{fig:sampling-scale} evaluates the performance on different scales of generated translations. 
The overall trends confirm the theoretical expectation that the performance improves when the scale increases. At the same time, the contribution of the gold translation drops when the scale increases, suggesting that with more generated translations, the gold translation provides less additional information. In addition, the performance of scale $\times 1$ and $\times 9$ have a gap of 0.75 s-BLEU, suggesting that the MT model requires sufficient samples from the DA model to match its distribution. In practice, we need to balance the performance gain and the training costs to decide on a suitable sampling scale.

\textbf{Observed Ratio.}
\label{subsec:observed-ratio}
Using the observed ratio ($\alpha$ in Eq. \ref{eq:posterior}), we can control the amount of information provided by the latent variable $z$. Such a ratio influences the quality of generated translations. As Figure \ref{fig:observe-ratio-a} shows, a higher observed ratio produces translations with a lower Deviation from the gold reference, which shows a monotonic descent curve. 
In comparison, the diversity of the generated translations shows a convex curve, which has low values when the observed ratio is small or big but high values in the middle. The diversity of the generated translations represents the degree of smoothness of the augmented dataset, which has a direct influence on the model performance. 

As Figure \ref{fig:observe-ratio-b} shows, the MT model obtains the best performance around the ratio of 0.4, where it has a balanced quality of Deviation and Diversity. When the ratio further increases, the performance goes down. Comparing the MT models trained with/without the gold translation, we see that the performance gap between the two settings is closing when the observed ratio is bigger than 0.6, where the generated translations have low Deviation from the gold translations.

The Diversity can be further enhanced by mixing the generated translations from different observed ratios. Therefore, instead of using a fixed ratio, we sample the ratio from a predefined Beta distribution. As Figure \ref{fig:observe-ratio-c} shows, we compare the performance on different Beta distributions. The performance on TED peaks at $Beta(1,1)$ but does not show a significant difference compared to the other two, while the performance on News peaks at $Beta(2,3)$, which has a unimodal distribution with an extremum between the ratio 0.3 and 0.4 and has a similar shape as the curve of Diversity in Figure \ref{fig:observe-ratio-a}. Compared to $Beta(2,2)$, which is also a unimodal distribution but with an extremum at the ratio 0.5, the performance with $Beta(2,3)$ is higher by 0.66 s-BLEU.

\textbf{Granularity of N-grams.}
The granularity of n-grams determines how much order information between tokens is observable through the latent $z$ (in comparison, the observed ratio determines how many tokens are observed). We evaluate different ranges of n-grams, where we sample n-grams according to a number uniformly sampled from the range. As Figure \ref{fig:observe-ngram} shows, the performance peaks at $[1,2]$ for TED and $[1,3]$ for News. However, the differences are relatively small, showing that the performance is not sensitive to the token order of the original reference. A possible reason may be that the DA model can reconstruct the order according to the semantic information provided by the source sentence.

\begin{table}[!t]
    \small\centering
    \setlength{\abovecaptionskip}{6pt}
    \setlength{\belowcaptionskip}{-9pt}
    \setlength{\tabcolsep}{2pt}
    \begin{tabular}{l|cc|c}
        \hline
        \multirow{2}{*}{\bf Method} & {\bf TED} & {\bf News} & {\bf Increase}  \\
         & s/d-BLEU & s/d-BLEU & s-BLEU  \\
        \hline
        G-Transformer (fnt.) & 25.20 / 27.94 & 25.12 / 27.02 & - \\
        + Source-side aug & 25.74 / 28.30 & 26.82 / 28.61 & +1.12 \\
        + Target-side aug & 26.59 / 29.20 & 28.06 / 29.83 & +2.17 \\
        + Both-side aug & 26.85 / 29.46 & 28.31 / 29.99 & +2.42 \\
        \hline
    \end{tabular}
    \caption{Source-side vs. target-side augmentations.}
    \label{tab:source-target-aug}
\end{table}

\subsection{Different Augmentation Methods}
\textbf{Source-side and Both-side Augmentation.}
We compare target-side augmentation with the source-side and both-side augmentations, by applying the DA model to the source and both sides. As Table \ref{tab:source-target-aug} shows, the source-side augmentation improves the baseline by 1.12 s-BLEU on average of TED and News but is still significantly lower than the target-side augmentation, which improves the baseline by 2.17 s-BLEU on average. Combining the generated data from both the source-side and target-side augmentations, we obtain an improvement of 2.42 s-BLEU on average, whereas the source-side augmented data further enhance the target-side augmentation by 0.25 s-BLEU on average. These results suggest that the DA model is effective for source-side augmentation but more significantly for target-side augmentation.

\begin{table}[t]
    \small
    \centering
    \begin{tabular}{lccc}
        \hline
        {\bf Method} & {\bf Dev} & {\bf Test} \\
        \hline
        Transformer (base) &  34.85 & 33.87 \\
        + T5 paraphraser $\diamondsuit$ & 34.01 & 33.10 \\
        + Target-side augmentation & {\bf 36.42}  & {\bf 35.42} \\
        \hline
    \end{tabular}
    \caption{Target-side augmentation vs paraphraser on sentence-level MT, evaluated on IWSLT14 German-English (De-En).  $\diamondsuit$ -- nucleus sampling with $p=0.95$.}
    \label{tab:paraphraser}
\end{table}

\textbf{Paraphrasing.}
Target-side augmentation augments the parallel data with new translations, which can be seen as paraphrases of the original gold translation. Such paraphrasing can also be achieved by external paraphrasers. We compare target-side augmentation with a pre-trained T5 paraphraser on a sentence-level MT task, using the settings described in Appendix \ref{subsec:paraphrases}.

As shown in Table \ref{tab:paraphraser}, the T5 paraphraser performs lower than the Transformer baseline on both the dev and test sets, while target-side augmentation outperforms the baseline by 1.57 and 1.55 on dev and test, respectively. The results demonstrate that a DA model is effective for sentence MT but a paraphraser may not, which can be because of missing translation information.

In particular, the generated paraphrases from the T5 paraphraser have a Diversity of 40.24, which is close to the Diversity of 37.30 from the DA model. However, when we compare the translations by calculating the perplexity (PPL) on the baseline Transformer, we get a PPL of 3.40 for the T5 paraphraser but 1.89 for the DA model. The results suggest that compared to an external paraphraser, the DA model generates translations more consistent with the distribution of the gold targets.

\subsection{Further Analysis}
\textbf{Size of The DA model.}
The condition on an observed translation simplifies the DA model for predicting the target. As a result, the generated translations are less sensitive to the capacity of the DA model. Results with different sizes of DA models confirm the hypothesis and suggest that the MT performance improves even with much smaller DA models. The details are in Appendix \ref{subsec:teacher-size}.

\textbf{Case Study.}
We list several word, phrase, and sentence cases of German-English translations, and two documents of English-German translations, demonstrating the diversity of the generated translations by the DA model. The details are shown in Appendix \ref{subsec:case-study}.

\section{Conclusion}
We investigated a target-side data augmentation method, which introduces a DA model to generate many possible translations and trains an MT model on these smoothed targets. Experiments show our target-side augmentation method reduces the effect of data sparsity issues, achieving strong improvement upon the baselines and new state-of-the-art results on News and Europarl. Analysis suggests that a balance between high Diversity and low Deviation is the key to the improvements. To our knowledge, we are the first to do target-side augmentation in the context of document-level MT.

\section*{Limitations}
Long documents, intuitively, have more possible translations than short documents, so a dynamic number of generated translations may be a better choice when augmenting the data, which balances the training cost and the performance gain. 
Another potential solution is to sample a few translations and force the MT model to match the dynamic distribution of the DA model using these translations as decoder input, similar to \citet{khayrallah2020simulated}. Such dynamic sampling and matching could potentially be used to increase training efficiency.
We do not investigate the solution in this paper and leave the exploration of this topic to future work.

Target-side augmentation can potentially be applied to other seq2seq tasks, where the data sparsity is a problem. Due to the limitation of space in a conference submission, we will leave investigations on other tasks for future work.

\section*{Acknowledgements}
We would like to thank the anonymous reviewers for their valuable feedback. This work is funded by the China Strategic Scientific and Technological Innovation Cooperation Project (grant No. SQ2022YFE020038) and the National Natural Science Foundation of China (grant NSFC No. 62161160339). Zhiyang Teng is partially supported by CAAI-Huawei MindSpore Open Fund (CAAIXSJLJJ-2021-046A). 


\bibliography{custom}
\bibliographystyle{acl_natbib}

\clearpage
\appendix

\section{G-Transformer}
\label{sec:gtrans}
G-Transformer \cite{bao2021g} has an encoder-decoder architecture, involving two types of multi-head attention. One is for global document, naming \emph{global attention}, while another is for local sentence, naming \emph{group attention}.

\textbf{Global Attention.}
The global attention is simply a normal multi-head attention, which attends to the whole document.
\begin{equation}
\small
\begin{split}
    args &= (Q, K, V), \\
    \text{GlobalAttn}(args) &= \text{softmax} \left( \frac{QK^T}{\sqrt{d_k}} \right) V,
\end{split}
\end{equation}
where matrix inputs $Q$, $K$, $V$ are query, key, and value for calculating the attention.

\textbf{Group Attention.}
The group attention differentiates the sentences in a document by assigning a group tag \cite{bao2021contextualized,bao2023general,bao2023gemini} to each sentence. The group tag is a number used to identify a specific sentence, which is allocated in the order of sentences, where the group tag for the first sentence is 1, second sentence 2, and so on. 

The group tag sequences are used to calculate an attention mask to avoid cross-sentential attention
\begin{equation}
\small
\begin{split}
    args &= (Q, K, V, G_Q, G_K), \\
    \text{GroupAttn}(args) &= \text{softmax} \left( \frac{QK^T}{\sqrt{d_k}} + M(G_Q, G_K) \right) V, \\
\end{split}
\end{equation}
where $G_Q$ and $G_K$ are group-tag sequences for query and key. The function $M(G_Q, G_K)$ calculates the attention mask that for a group tag in $G_Q$ and a group tag in $G_K$, it returns a big negative number if the two tags are different, otherwise it returns 0. 

\textbf{Combined Attention}
The two multi-head attentions are combined using a gate-sum module
\begin{equation}
\small
\begin{split}
    H_L &= \text{GroupMHA}(Q, K, V, G_Q, G_K), \\
    H_G &= \text{GlobalMHA}(Q, K, V), \\
    g &= \text{sigmoid}([H_L, H_G]W + b), \\    
    H &= H_L \odot g + H_G \odot (1 - g), \\
\end{split}
\label{eq:combineattn}
\end{equation}
where $W$ and $b$ are trainable parameters, and $\odot$ denotes element-wise multiplication.

G-Transformer uses group attention on low layers and combined attention on top 2 layers.

\section{Datasets and Metrics}
\subsection{Datasets}
\label{subsec:datasets}
The three benchmark datasets are as follows.

\textbf{TED} is a corpus from IWSLT2017, which contains the transcriptions of TED talks that each talk corresponds to a document. The sentences in source and target documents are aligned for translation. We use {\it tst2016-2017} for testing and the rest for development.

\textbf{News} is a corpus mainly from News Commentary v11, where the sentences are also aligned between the source and target documents. We use {\it newstest2016} for testing and {\it newstest2015} for development. In addition, we use {\it newstest2021} from WMT21 \cite{farhad2021findings}, which has three references for each source, to evaluate the quality of the estimation of data distribution.

\textbf{Europarl} is a corpus extracted from Europarl v7, where the train, development, and test sets are randomly split.

We pre-process the data by tokenizing and truecasing the sentences using MOSES tools \cite{koehn2007moses}, followed with a BPE \cite{sennrich2016neural} of 30000 merging operations.

\subsection{Metrics}
\label{subsec:metrics}
The sentence-level BLEU score (s-BLEU) and document-level BLEU score (d-BLEU) are described as follows.

\textbf{s-BLEU} is calculated over sentence pairs between the source and target document, which is basically the same with the BLEU scores \cite{papineni2002bleu} for sentence NMT models.

\textbf{d-BLEU} is calculated over document pairs, taking each document as a whole word sequence and computing the BLEU scores between the source and target sequences.

\begin{figure}[t]
    \centering
    \includegraphics[trim={0 12pt 0 12pt},clip,width=0.8\linewidth]{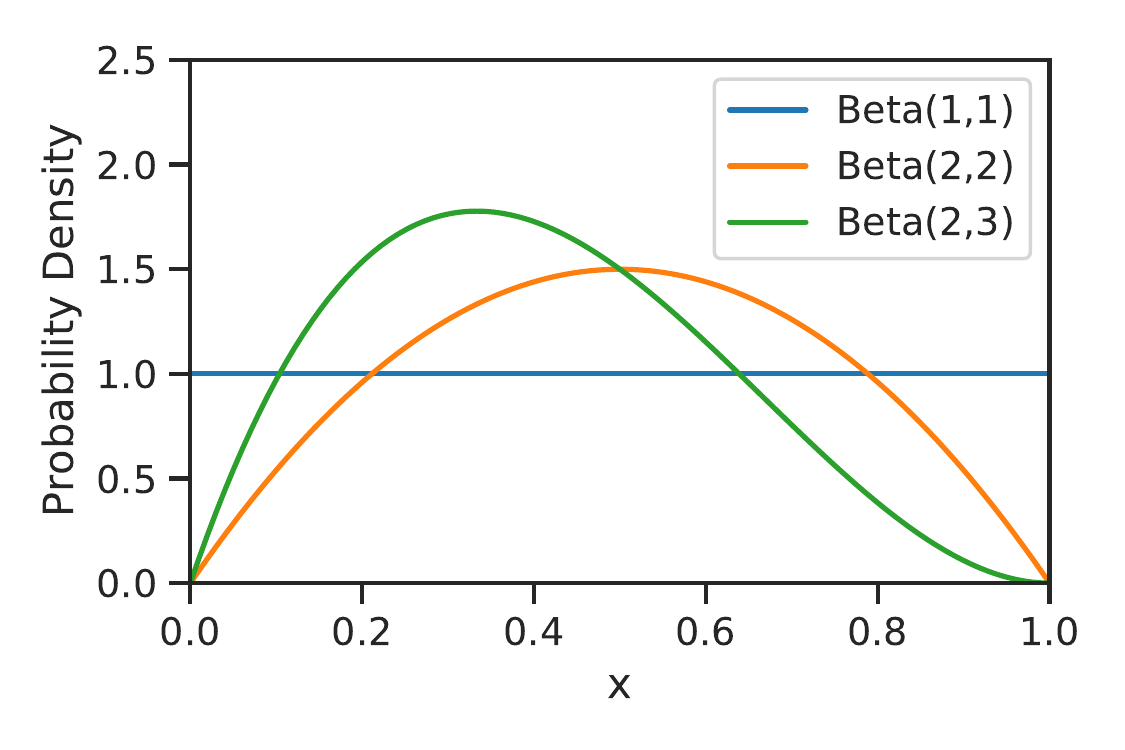}
    \caption{The probability density function of $Beta(a,b)$ distributions.}
    \label{fig:beta-distrib}
\end{figure}

\begin{table*}[t]
    \small
    \centering
    \begin{tabular}{l|cc|ccc|c}
        \hline
        {\bf Teacher Size} & {\bf Hyperparameters} & {\bf Params} & {\bf TED} & {\bf News} & {\bf Europarl} & {\bf Average}  \\
        \hline
        Base & 6 layers, 8 heads, hidden 512, FFN 2048 & 69M & 26.59 & 28.06 & 32.85 & 29.17 \\
        Small & 6 layers, {\bf 4} heads, hidden 512, FFN {\bf 1024} & 56M & 26.64 & 28.23 & 32.42 & 29.10 \\
        Tiny & 6 layers, {\bf 4} heads, hidden {\bf 256}, FFN {\bf 1024} & 21M & 26.73 & 28.08 & 32.30 & 29.04 \\
        \hline
    \end{tabular}
    \caption{Impact of the size of the DA model, trained on G-Transformer (fnt.) and evaluated in \emph{s-BLEU}.}
    \label{tab:teacher-size}
\end{table*}

For \emph{analysis}, we measure the Deviation and Diversity of generated translations.

\textbf{Deviation} is simply defined as the distance to perfect s-BLEU score
\begin{equation}
  \text{Deviation}(\hat{y}, y) = 100 - \text{s-BLEU}(\hat{y}, y), \\
  \label{eq:deviation}
\end{equation}
where $\hat{y}$ is a generated translation and $y$ is the reference translation.

\textbf{Diversity} is calculated by averaging the deviation scores among the generated translations
\begin{equation}
  \text{Diversity}(\mathcal{\hat{Y}}) = \frac{\sum_{i=1}^{M} \sum_{j=i+1}^{M} \text{Deviation}(\hat{y}_i, \hat{y}_j)}{M(M-1)/2}, \\
  \label{eq:diversity}
\end{equation}
where $\mathcal{\hat{Y}}$ is a set of generated translations, containing $M$ elements. The metric is similar to a diversity metric in \citet{he2018sequence} beside that we use s-BLEU for basic measure.

\subsection{Training Settings}
\label{subsec:training-settings}
We use a base model for all the baselines, where the models have around 60M parameters. We adjust several hyper-parameters of the default setting to better suit the augmented data. First, we extend the maximum length of the model from 512 to 1024. Next, we change the dropout from 0.3 to 0.1 for Europarl but keep the dropout of 0.3 for News and TED. Last, we reduce the patience of training the DA model from 10 to 5 for TED and News, and from 10 to 2 for Europarl, so that the training process could be accelerated.

Running with the new settings on 4 Tesla V100 GPUs, the G-Transformer (fnt.) baseline takes 2 hours to train on TED, 2.5 hours on News, and 13 hours on Europarl. After augmenting the data 9 times, the training of G-Transformer (fnt.) for the MT model costs 10, 16, and 49 hours on TED, News, and Europarl, respectively. In comparison, the training for the DA model costs 5, 8.5, and 25 hours on TED, News, and Europarl, respectively.

\textbf{Beta Distributions.}
\label{subsec:beta}
We use a Beta distribution to sample the observed ratio, where we consider three basic candidates including $Beta(1,1)$, $Beta(2,2)$, and $Beta(2,3)$ as Figure \ref{fig:beta-distrib} displays.

We decide on the choice by comparing the figure to the Diversity curve shown in Figure \ref{fig:observe-ratio-a}, where we can see that $Beta(2,3)$ has the best match with the Diversity curve of the generated translations. Our further analysis in Section \ref{subsec:observed-ratio} confirms that $Beta(2,3)$ provides a balanced performance on TED and News.

\section{More Analysis}
\subsection{Multi-reference Evaluation}
\label{subsec:estimate-distrib}
As more direct evidence that a DA model with a posterior distribution estimates $P_{data}(y|x_i)$ more accurately than that with a prior distribution, we evaluate the perplexity (PPL) on a multiple-reference dataset \emph{newstest2021}, which contains 67 documents and 1002 source sentences, each with 3 translations. We cross-validate the translations by using one as an observed translation and the other two as test translations. Using Eq. \ref{eq:posterior-approx}, we approximate the posterior probability by sampling the latent $z$ sufficient times (e.g., 100).

\subsection{Size of The DA Model}
\label{subsec:teacher-size}
The posterior distribution simplifies the translation task for the DA model since the input latent $z$ contains much information about the target. As a result, the DA model is less sensitive to the capacity of the model. We evaluate target-side augmentation with different sizes of DA models. The results are shown in Table \ref{tab:teacher-size}. 
The performance on TED and News does not show a significant difference when we reduce the number of parameters from 69M to 21M. On bigger Europarl, the performance drops by 0.55 s-BLEU but still outperforms the baseline G-Transformer (fnt.) by 0.37 s-BLEU, suggesting that the DA model provides additional value even when its capacity is much lower than the MT model.

\begin{table*}[t]
    \centering\small
    \setlength{\tabcolsep}{6pt}
    \renewcommand{\arraystretch}{1.8}
    \begin{tabular}{@{}lp{3cm}p{3cm}p{7.5cm}@{}}
        \hline
        {\bf Level} & {\bf Source} & {\bf Target} & {\bf Generated Translations} \\
        \hline
        \multirow{3}{*}{Word} & herauszufinden & identify & find out, figure out, find, learn, look out, see  \\
        & unglaublich & incredibly & unbelievable, amazingly, extremely, highly, remarkably \\
        & überzeugt & convinced & persuaded, believed, pretty sure  \\
        \hline
        \multirow{3}{*}{Phrase}  & halten diese Einschränkungen für sinnvoll & accept such limits as reasonable & 1) {\it consider these restrictions useful} \newline 2) {\it regard such restrictions as reasonable} \newline 3) {\it take these constraints as certain} \\
        & passiv bewegte ohren sobald der kopf etwas tut . & ears that move passively when the head goes . & 1) {\it ears moving passively} when the head does something . \newline 2) {\it passively moving ears} once the head goes . \newline 3) {\it passive ears that move} when your head does something . \\
        & ein aus holz und stoff gebautes objekt ist , mit eingebauten bewegungen , um euch glauben zu lassen , sie sei lebendig . & an object constructed out of wood and cloth with movement built into it to persuade you to believe that it has life & 1) {\it an object made out of} wood and cloth , with movement built in to persuade you to believe that {\it it 's alive} . \newline 2) an object built out of wood and cloth {\it with movement to perpetuate you to believe it 's alive} . \newline 3) {\it a wooden and cloth object} with movement built in to make you believe that it 's alive . \\
        \hline
        \multirow{3}{*}{Sentence} & sie lebt nur dann wenn man sie dazu bringt . & it only lives because you make it . & 1) it only lives {\it when you get it to do} . \newline 2) it lives {\it only as you make it} . \newline 3) it only lives {\it because you get them to do it} .  \\
        & in jedem moment auf der bühne rackert sich die puppe ab . & so every moment it 's on the stage , it 's making the struggle . & 1) {\it at every moment on the stage} , it 's making the struggle of puppet . \newline 2) every moment on the stage {\it it reckers down the puppet} . \newline 3) so every moment it 's on the stage , {\it the puppet is racking off} . \\
        & er demonstriert anhand einer schockierenden geschichte von der toxinbelastung auf einem japanischen fischmarkt , wie gifte den weg vom anfang der ozeanischen nahrungskette bis in unseren körper finden . & he shows how toxins at the bottom of the ocean food chain find their way into our bodies , with a shocking story of toxic contamination from a japanese fish market . & 1) {\it he demos through a shocking story of toxic burden} on a japanese fish market , {\it how poisoning their way from the beginning of the ocean food chain into our bodies} . \newline 2) he demos through a shocking story of toxin impact on a japanese fish market , {\it how poised the way from the ocean food chain to our bodies} . \newline 3) he demos through a shocking story of toxin contamination at a japanese fish market , {\it with how toxins find the way from the beginning of the ocean food chain to our bodies} . \\
        \hline
    \end{tabular}
    \caption{Translations generated by the DA model on IWSLT14 German-English.}
    \label{tab:paraphrases}
\end{table*}

\subsection{Paraphrasing Settings}
\label{subsec:paraphrases}
We use the T5 paraphraser \footnote{https://huggingface.co/Vamsi/T5\_Paraphrase\_Paws}, created by fine-tuning T5 \cite{raffel2020exploring} on English paraphrases \cite{zhang2019paws}, as a representative to make a comparative study. 
Given that the T5 paraphraser is trained in English and works at the sentence level, we translate the documents sentence-by-sentence and evaluate the methods on MT benchmark IWSLT14 German-English. For each target sentence, we sample 6 paraphrases by running nucleus sampling \cite{holtzman2019curious} with the T5 paraphraser. For target-side augmentation, we generate 6 translations for each source sentence without using the document context.
It is worth noting that different from the previous paraphrasing augmentation method \cite{khayrallah2020simulated}, where the MT model learns the paraphraser's distribution directly, we use sampled text output to train the MT models.

\begin{table*}[t]
    \centering\small
    \setlength{\tabcolsep}{6pt}
    \renewcommand{\arraystretch}{1.5}
    \begin{tabular}{p{15.5cm}}
        \hline
        \textbf{Source:} Elton John and Russian President Vladimir Putin to meet to discuss gay rights in 2003, Mikhail Khodorkovsky, Russia 's wealthiest man, was arrested at gunpoint on a Siberian runway. having openly challenged President Vladimir Putin, Khodorkovsky was convicted, his oil company, Yukos, seized and his pro democracy efforts curtailed. \\
        \textbf{Target:} Elton John und der russische Präsident Vladimir Putin treffen sich, um Rechte der Schwulen zu diskutieren Mikhail Khodorkovsky, {\it Russlands reichster Mann}, wurde auf einem sibirischen Rollfeld mit Waffengewalt verhaftet. nachdem er Präsident Vladimir Putin offen herausgefordert hatte, wurde Khodorkovsky verurteilt, sein Ölunternehmen Yukos beschlagnahmt und {\it seine demokratischen Bemühungen unterbunden}. \\
        \textbf{Baseline:} Elton John und der russische Präsident Wladimir Putin müssen sich treffen, um über Homosexuelle zu diskutieren im Jahr 2003 wurde Michail Chodorkowski, {\it der reichste Mann Russlands}, an einer sibirischen Stichwahl verhaftet. nachdem er Präsident Wladimir Putin offen in Frage gestellt hatte, wurde Chodorkowski verurteilt, seine Ölgesellschaft Yukos, beschlagnahmt und {\it seine Anstrengungen zur Demokratie beschnitten}. \\
        \textbf{Ours:} Elton John und der russische Präsident Wladimir Putin treffen sich, um über Homosexuellenrechte zu diskutieren 2003 wurde Michail Chodorkowski, {\it Russlands reichster Mann}, auf einer sibirischen Stichwahl verhaftet. nachdem er Präsident Wladimir Putin offen in Frage gestellt hatte, wurde Chodorkowski verurteilt, seine Ölgesellschaft Yukos erobert und {\it seine Bemühungen zur Demokratie eingeschränkt}. \\
        \hline
        \textbf{Source:} the Upper Bavarian district of Ramsau bei Berchtesgaden is Germany 's first "Mountaineers 'Village". the village of 1,800 inhabitants in the Berchtesgaden National Park received the award for "gentle Tourism" from the hand of the Vice President of the German Alpine Association, Ludwig Wucherpfenning, on Wednesday. there are already 20 "Mountaineers' Villages" in Austria. in our neighbouring country, the local Alpine Association is responsible for awarding the distinction. a "Mountaineers 'Village" is permitted to have a maximum of 2,500 residents. at least one fifth of its area must be designated as a protected area. \\
        \textbf{Target:} die oberbayerische Gemeinde Ramsau bei Berchtesgaden ist Deutschlands erstes "Bergsteigerdorf". {\it aus der Hand des Vizepräsidenten} beim Deutschen Alpenverein, Ludwig Wucherpfennig, erhielt das 1800-Einwohner-Dorf im Nationalpark Berchtesgaden am Mittwoch die Auszeichnung für sanften Tourismus. in Österreich gibt es bereits 20 "Bergsteigerdörfer". im Nachbarland ist der dortige Alpenverein für die Vergabe der Auszeichnung zuständig. ein "Bergsteigerdorf" darf höchstens 2500 Einwohner haben. mindestens ein Fünftel seiner Fläche muss als Schutzgebiet ausgewiesen sein. \\
        \textbf{Baseline:} der Upper Bavarian Distrikt Ramsau und Berchtesgaden ist Deutschlands erste,, Mountaineers 'Village ". das Dorf von 1.800 Einwohnern im Berchtesgaden National Park erhielt den Preis für den,, sanften Tourismus" {\it von der Hand des Vizevorsitzenden} der Deutschen Alpine Association, Ludwig Wucherpfing am Mittwoch. in Österreich gibt es bereits 20,, Mounineers' Villages ". in unserem Nachbarland ist die lokale Alpine Association dafür verantwortlich, diese Unterscheidung zu vergeben. ein,, Mountagiers 'Village" darf ein Maximum von 2.500 Einwohnern haben. mindestens ein Fünftel der Gegend muss als geschütztes Gebiet ausgewiesen werden. \\
        \textbf{Ours:} der Upper Bavaristische Bezirk Ramsau bei Berchtesgaden ist Deutschlands erstes "Mountaineers 'Village". das Dorf mit 1.800 Einwohnern im Berchtesgaden National Park erhielt am Mittwoch den Preis für "sanften Tourismus" {\it aus der Hand des Vizepräsidenten} der deutschen Alpine Association, Ludwig Wucherenning. in Österreich gibt es bereits 20 "Mountaineers' Villages". in unserem Nachbarland ist die lokale Alpine Association dafür verantwortlich, diese Unterscheidung zu vergeben. ein "Mountaineers 'Village" darf ein Maximum von 2.500 Einwohnern haben. mindestens ein Fünftel seines Gebietes muss als geschützte Gegend bezeichnet werden. \\
        \hline
    \end{tabular}
    \caption{Comparison of the document-level translations from G-Transformer (fnt.) baseline and target-side augmentation, evaluated on {\it News} English-German.}
    \label{tab:doc-trans}
\end{table*}

\subsection{Case Study}
\label{subsec:case-study}
Our case study demonstrates that the DA model generates diverse translations at word, phrase, and sentence levels. Several cases for German-English translation are listed in Table \ref{tab:paraphrases}. 

We further list two document-level translations, through which we can have a direct sense of how target-side augmentation improves MT performance, as Table \ref{tab:doc-trans} shows.

\end{document}